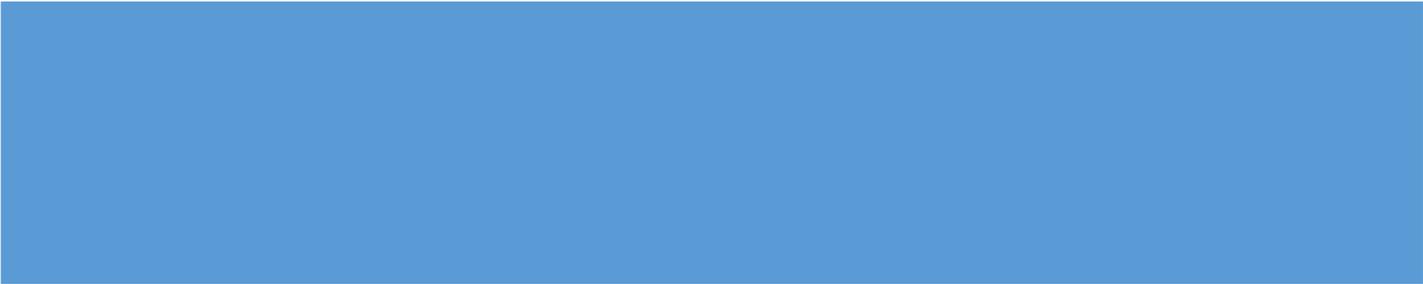

# LITERATURE REVIEW
## TEXTURELESS OBJECT RECOGNITION

Frincy Clement | Kirtan Shah | Dhara Pancholi

frincy@ualberta.ca | knshah@ualberta.ca | dhara@ualberta.ca
UNIVERSITY OF ALBERTA

# Contents

# 1. TEXTURELESS OBJECT RECOGNITION

## I. INTRODUCTION

In recent years, a lot of labor-intensive activities in manufacturing sector has been automated using robots. In industrial assembly line, a robotic arm is used to pick up small objects and place it in its location. These industrial objects are mostly texture-less and are different from textured objects in terms of lack of discriminative features. Often, they are smaller in size and can be similar to each other. In real world, they are present along with multiple objects of different classes in different scene configurations, which makes the recognition even more challenging.

In the early 2000s, detection and recognition of the texture-less objects seemed to be a daunting issue, even for the state-of-art detectors at that time, like SIFT (Scale Invariant Feature Transform) and SURF (Speeded UP Robust Features). This was due to their dependence on the high-featured, texture-rich informative regions which were sparse in the texture-less occurrences. Ultimately, this led to several works which contributed for the detection and recognition of texture-less objects, which can be thus classified into three categories: View-based, Feature-based, and Shape-based.

## II. RELATED WORK

### a) View-based

View-based methods work by comparing the object of interest (OOI) and the pre-calculated 2D views of that object, denoted by aspect graphs. This method can provide a rough estimate of the pose of the object, but it suffers from computational load of searching a large spatial domain. To overcome this issue, Cyr and Kimia [3] proposed a method that grouped similar OOI together under a single class and similarity scores were calculated to perform recognition. Eggert *et al.* [1], Ulrich *et al.* [8] and Steger *et al.* [2] introduced various improvements on the aspect graphs and similarity metric. This approach, however was not widely used because of its high complexity.

### b) Feature based method

Tombari *et al.* [9] devised a method called BOLD in which they attempt to tackle the issue of object detection under clutter. In this method, neighbouring line segments were aggregated to form a feature representation of the object. These line segments are invariant to rotation, translation as well as scaling.

A major challenge of this approach was the selection of an optimal value of 'k'. To overcome this challenge, they introduced an approach called multiple bunch where a bunch of 'k's were used instead of a single threshold. For a single 'k' approach, the optimum value of k was 10, which comprised of good descriptiveness and very little noise. However, there was a significant improvement in the performance when multiple 'k's were used.

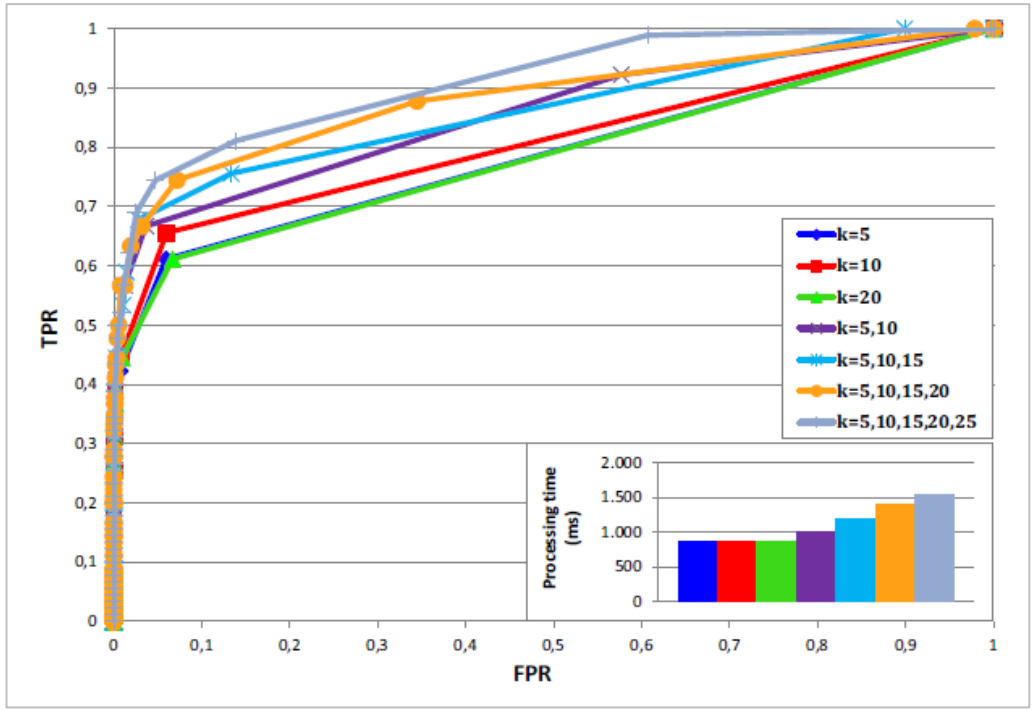

**Figure: BOLD descriptions employing single vs. multiple bunches with different k values**
Source: Tombari *et al.* [9]

A limitation of BOLD is that it fails if there is clutter in the line segment selected for feature description.

Intended for object recognition and tracking using an RGB-D sensor, Jiang *et al.* [18] proposed a method that detects and recognises small objects. The image is divided into various parts using image segmentation. Then it is sent to a classifier as an input which classifies the object using the features extracted in the earlier stages.

Hodan *et al.* [13] proposed a technique for detecting textureless objects using AR, which used a purely edge based detector which gave better results than canny edge detector. The results showed this method is 5 times faster than canny and efficient towards the object detection as it searched edgelet in stripes instead of wedges. When the binary map of edges is provided, this method takes the edglets of equal fix lengths and they are represented by their midpoints and orientation, which helps in tracking the orientation. Two images are matched using their distance transformation in orientation which proves the presence of that object in image even if more than one same object is present.

While the above-mentioned methods do a commendable job of detecting texture-less objects, they were computationally expensive. In 2016 Jacob *et al.* [19] introduced BORDER which is an acronym for Bounded Oriented Rectangle Descriptors for Enclosed Region, in which they use a rectangle to encapsulate the object so that minimum outliers are detected. It outperformed the previous state-of-the-art descriptors BOLD and Line2D when compared for images having occlusion and clutter. It divides the line segment into small equal fragments called linelets. Then, these linelets are established into the oriented rectangles to cover the region inside the object. Further to diminish the effect of the occlusion, the rectangle is rotated at different angle to form multiple rectangle layers. It follows the structure of the SIFT technique which contains three steps: Detection, Description and Matching.

Counting the number of texture-less objects present in a scene was still a challenge until Verma *et al.* [21] proposed a method which counts the objects using shape and colour feature. First, the said features were extracted during pre-processing using morphological boundary extraction and segmentation via mean hue value (colour features) and Hu-moments (shape features). Next, SVM, kNN, NN and tree bagging were applied for the classification of objects. Through experiments, they observed that tree bagging has the best accuracy of all the classifiers. Finally, the classified objects were counted by drawing bounding boxes around them.

To improve accuracy, Thoduka *et al.* [23] in 2016 came up with an approach which used RGB-D based features to recognize objects in the industrial assembly line such as nuts, bolts, bearing large and small black profiles etc. From the RGB-D images captured using a live 3D camera, they extracted the 3D point cloud of all objects in the dataset. Four feature descriptors such as bounding box, median/mean color, size of point cloud and circularity were calculated and passed to an SVM classifier for recognition. They concluded that adding more features does not result in better accuracy, rather made the model overfit to training set. The approach performed efficiently on most of the objects in dataset but they had poor performance recognizing smaller objects.

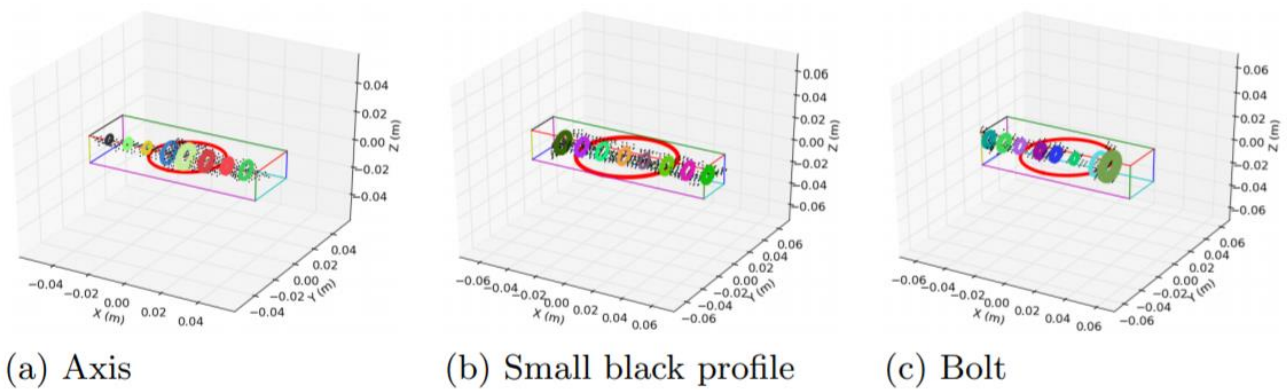

**Figure: Illustration of features calculated on the 3D point cloud of objects in dataset**
Source: Thoduka *et al.* [23]

Following upon his scholarly work, Hodan *et al.* [25] improved upon his previous research by presenting a new dataset in 2017, "T-less" publicly for better understanding of 3D object from different dimensions. This dataset contains industrial objects which have no texture while having feature similarities of shape/size which makes it quite hard to differentiate.

There are several datasets mentioned in the paper but among them T-less proved to be useful as they have good room for rectification. Reason for this is: they are large in numbers, training images are taken in authorized condition, objects are seen with high clutter and occlusion, images are captured through three different sensors and it provides two 3D models for each object.

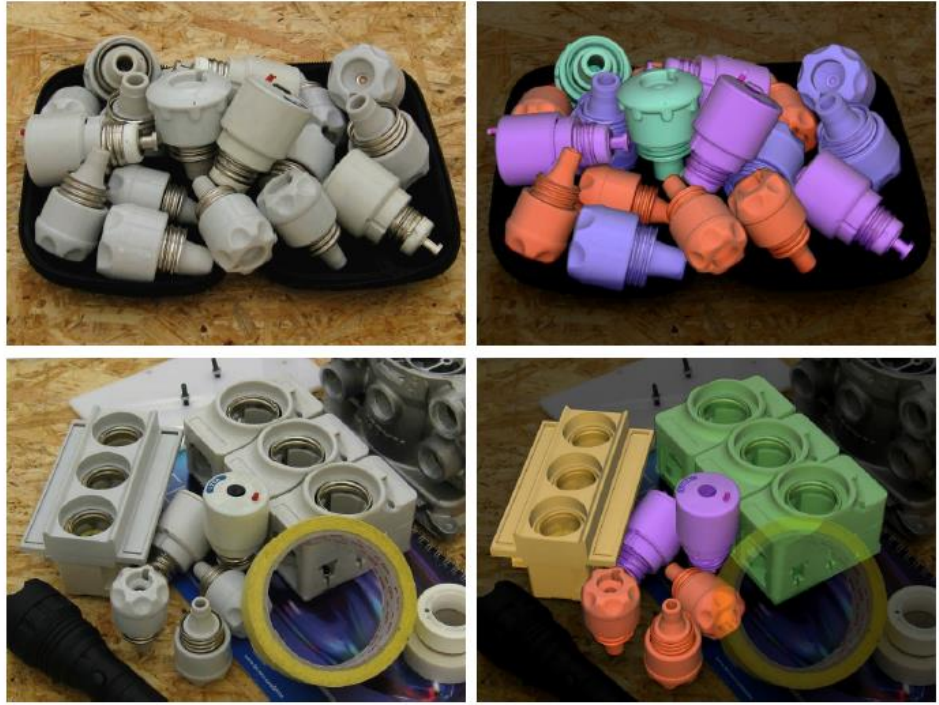

**Figure: TLess Dataset by Hodan *et al*. [25]**

To overcome the short comings of previous methods, BIND [24] was introduced during 2017. It is a detector that uses binary nets having multiple layer for accurate detection of texture less object. Initially, it encapsulates the object by net, which is then divided into equal fixed blocks.

These blocks give details of the object body in the form of binary bits. The main advantage of using BIND is, its invariant towards properties like scaling, rotation, translation etc. The matching of the template and actual scene is done through bitwise AND & OR operator.

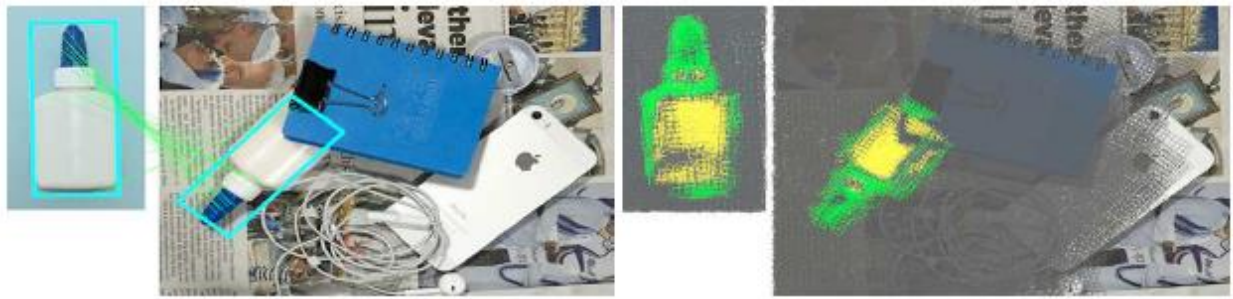

**Figure: Key point matches along with the object detection bounding box results using BIND are presented on the left, and the net matches with green blocks indicating oriented blocks, yellow signifying internal homogeneous blocks, and gray implying background/no-match blocks are displayed on the right**
Source: Chan *et al.* [24]

To improve the speed of object detection, in 2018 Hancheng *et al.* [37] introduced a faster approach to detect texture-less objects even in the cluttered background and transformations. It is a two-step

method - first it compresses all the angular orientations nearby then forms an orientation compressing map which is used to know the object location and in next step it computes the similarities between pixels using the Discriminative Regional weights. Each orientation is described by a binary string. Here the canny edge detector is used for edge detection. Since the template and input have used bitwise AND operation for encoding and compressing, it will result in a faster detection.

Fang *et al.* [33] deduced that for object recognition, the background of an image affects the performance of neural networks, and hence came up with a method called DOG, built upon CNN which effectively removed the background of an input image for improving the efficiency of the classifier. Fully Connected Layers (FCNs) such as U-Net and Tiramisu were applied to the image to observe their performance. Tiramisu was used for DOG.

c) Shape Based Approach

In the study of texture-less object detection, template matching is a popular method dating back as early as the 2000s in lieu of the fact that it provides high mAP and close to real-time detection. But, up until now, templates were generated using one modality only (depth or colour or point cloud or 2D). However, in 2011 Hinterstoisser *et al.* [6] put forth the idea of multi-modal template matching. The template comprises of both the image cue (colour information) as well as depth cue (3D) information. The method can detect objects in real-time as well as under heavy clutter.

The method extracts gradients from colour images as they are more robust to illumination changes compared to gradients extracted from grayscale images. Also, 3D surface normal is calculated which are generally found on the inside of the object. The gradients and surface-normals are summed to obtain a multi-modal template. For new objects, one only needs to extract and store multi-modal features which are instantaneous. Hence, it is real-time object recognition.

A caveat of the above approach is that it is not invariant to translation, rotation, or scaling.

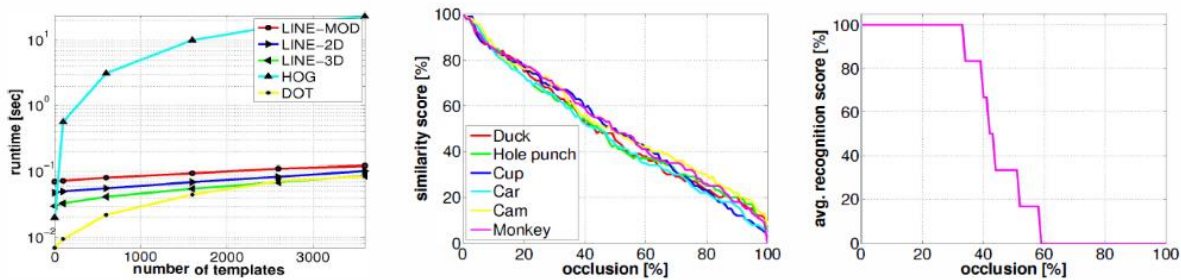

**Figure: Comparison between LINEMOD, LINE-2D, LINE-3D, HOG and DOT**
Source: Hinterstoisser *et al.* [6]

In the following year, Hinterstoisser *et al*. [7] overcame the afore-mentioned issue by considering all the gradient orientations for a template instead of just the dominant ones. Due to this, the method was more robust to translations. First, a similarity measure is calculated to understand the importance of each aspect of an image. Also, depth data can be integrated with the gradients to improve the robustness of the technique. A similarity measure is calculated for an orientation that searches the neighbourhood of an input image for templates with similar orientations. While it does suppress the caveat of the above approach to a certain extent, it is still not invariant to scaling.

In 2015 Hodan *et al.* [13] came up with a very efficient way of template matching, resolving a lot of its computational complexity in performing searches of each sliding window with each stored template. By calculating an objectness score for each image window, which is a measure of presence of an object, they reduce the number of image windows by removing those which do not pass the threshold score. For each of these selected windows, a subset of candidate templates with highest probabilities are identified using a voting system based on hash tables. Now, this reduced number of templates are verified with respect to a set of five different features for each selected window. After applying non-maximal suppression, the irrelevant bounding boxes with large overlap to those with highest score are removed. The remaining detections are the result for object recognition. This method resulted in real time performance with low false rates and multiple object detection.

In 2018, an edge-based Hierarchical Template Matching algorithm was proposed by Tsai *et al.* [29] based upon Line2D algorithm, which gave an efficient method for real-time object detection and recognition.

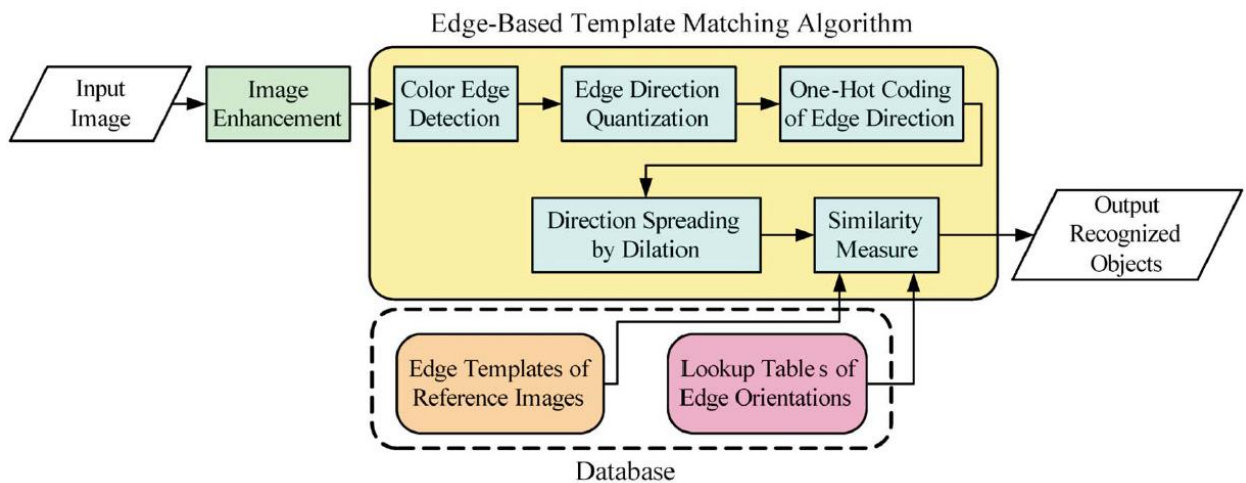

**Block Diagram of the Texture-less Object Detection and Recognition based on Edge-based Template matching Algorithm**
Source: Tsai *et al.* [29]

Edges are identified as the most stable feature for a texture-less object. After image enhancement, the input image goes through the edge-based template matching algorithm. It detects the color edge by finding the one with maximum magnitude across all color channels. These detected edges are then quantized into one of 8 bins and one-hot encoded for computational efficiency. The one-hot encoded edge orientations are then spread using dilation.

The algorithm then checks for similarity with the precomputed database which has edge templates of reference images and lookup tables of edge orientations, which will help in efficient matching and calculation of similarity score.

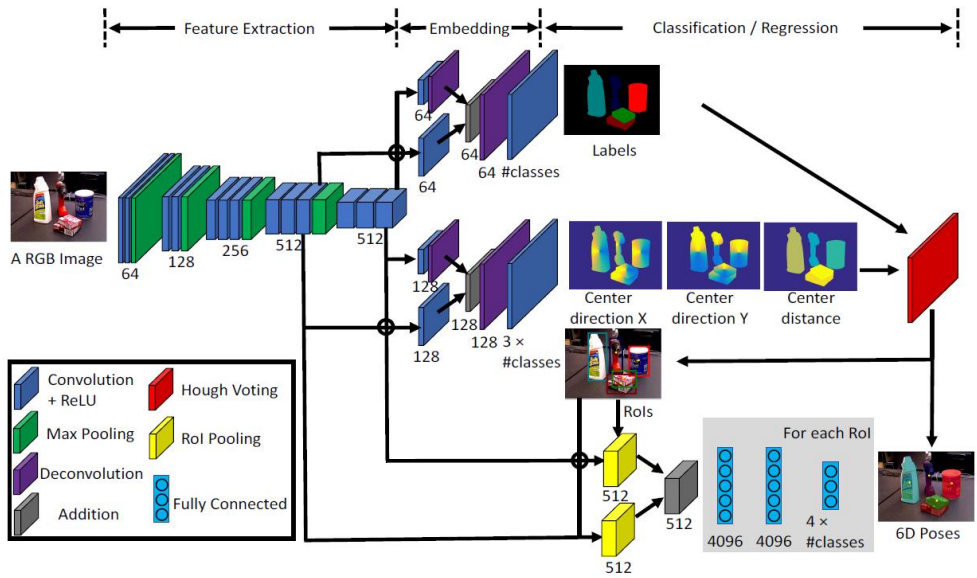

**Figure: Architecture of PoseCNN**
Source: Xiang *et al*. [27]

2018 saw many approaches for texture-less object detection. PoseCNN (current state-of-the-art), an approach by Xiang *et al.,* [27] is a convolutional network for predicting object pose with varied orientations. It predicts object's translation matrix by means of calculating object's distance with respect to the camera. Also, they calculate the object's rotation matrix.

PoseCNN performs three main tasks – i) Firstly, it gives a semantic label to all the objects in the dataset. ii) Next, it calculates the distance between the object center to the center of the image. Since the camera intrinsic is known, 3D translation poses can be estimated. iii) Finally, 3D rotation R is predicted by using a quaternion representation.

Midway through 2019, Park *et al.* [35] proposed a method that takes depth image as an input and a region of interest (ROI) is generated. ROI is a patch where the network believes there to be an object/s. The ROI is then reduced to 14x14 having 256 filters, further descriptors are extracted from it. This ROI is then compared with an ROI from the template that is pre-calculated for image segmentation and pose estimation.

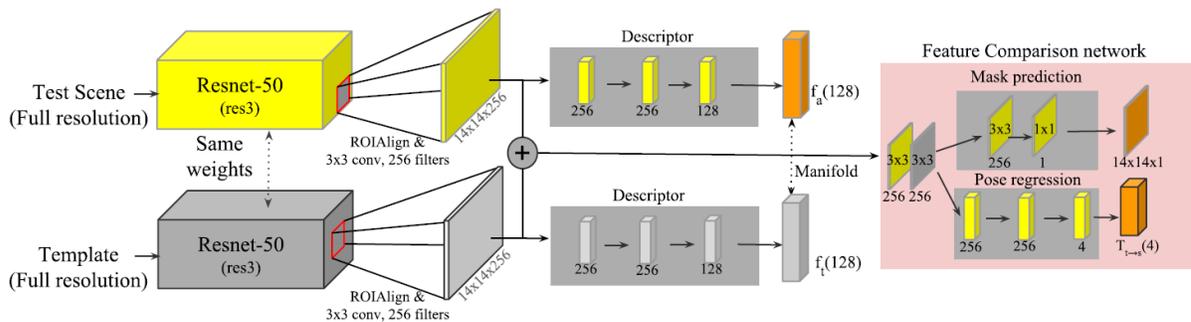

**Figure: Architecture of Multi-Task Template Matching**
Source: Park *et al.* [35]

## III. INPUT PREPROCESSING

Before an image is subjected to a classifier, it must be pre-processed for efficient classification. In this section we will discuss some prominent pre-processing techniques.

A paper proposed by Al-amri *et al.* [5] in 2010 compares various edge detectors. One of the most important uses of edge detection technique is for image segmentation. An image is divided into various regions which can further be classified as foreground and background. An edge is generally associated with a significant change in the intensity of the image or some discontinuity. Discontinuity in a grayscale image can be of three types - point, line, edge, and gradient.

- Sobel operator uses two masks to find the gradient magnitude of the image from the original image.
- Robert's technique operates on the fact that high spatial frequency relates to edges in an image. Hence, it performs a quick, 2D spatial gradient operation on the input image.
- Laplacian technique performs a second-order derivation on an input image f(x,y). It provides an inference of whether the pixel is dark (background) or light (foreground).
- Prewitt edge detector finds the orientation and magnitude of an edge. It makes use of kernel Gx (Vertical) and Gy (Horizontal) and summation of it provides the intensity of the gradient.
- Kirsch Operator uses four templates instead of two to calculate the edges.
- The canny technique attempts to maximize the number edges by minimizing the error rate. First, it convolves the image with a gaussian function to smooth the image. Next, it calculates the first difference of the gradient to obtain the magnitude and edge strength. Finally, thresholded NMS is applied on top of it.

The study revealed that Kirsch, edge maximization technique and Prewitt are the optimal edge detectors.

**Figure: Prewitt = Sobel where c=1**
Source: Al-amri et al. [5]

Following upon the previous work, Akanksha et al. [11] proposed another work that represents importance of segmentation for object recognition. Segmentation is defined as separating out the regions with similar textures, intensity etc. with respect to background. Then the objects with same class are kept together, so as a result, the objects are separated according to the classes. There are several image segmenting techniques like Otsu's method and K-means clustering. In this paper, they compared several such methods to find the best segmentation technique.

There are three kind of approaches as shown in the figure:

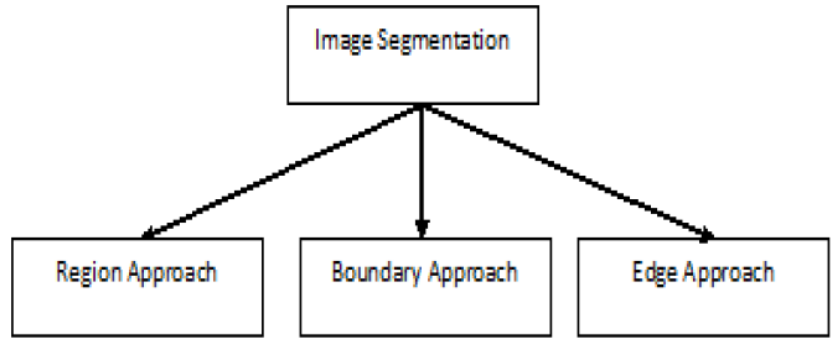

**Figure: Different types of image segmentation**
Source: Akanksha *et al*. [11]

Image segmentation method based on region includes: Region growing method, Region splitting & merging method and Clustering method. Segmentation based on edge detection includes: Ray histogram technique, Gradient based method, first order derivative method (Prewitt operator, Sobel operator, Canny operator) and Second order derivative method which has Laplacian operator and Zero crossing. Other type of segmentation is done through thresholding methods like Global thresholding and Local thresholding.

The main aim of the paper introduced by Aziz *et al*. [14] is to show different image pre-processing techniques that uses high and low pass filters, which make use of Fourier and wavelet transformation. Low pass filters are for smoothening the images whereas high pass filters are for sharpening the images. They proved that high and low pass filters are efficient for image transformation of noisy images. The filter function in the diagram depicts low and high pass filter function. There are several low pass filters: Ideal low pass filters, Gaussian low pass filters and Butterworth low pass filters.

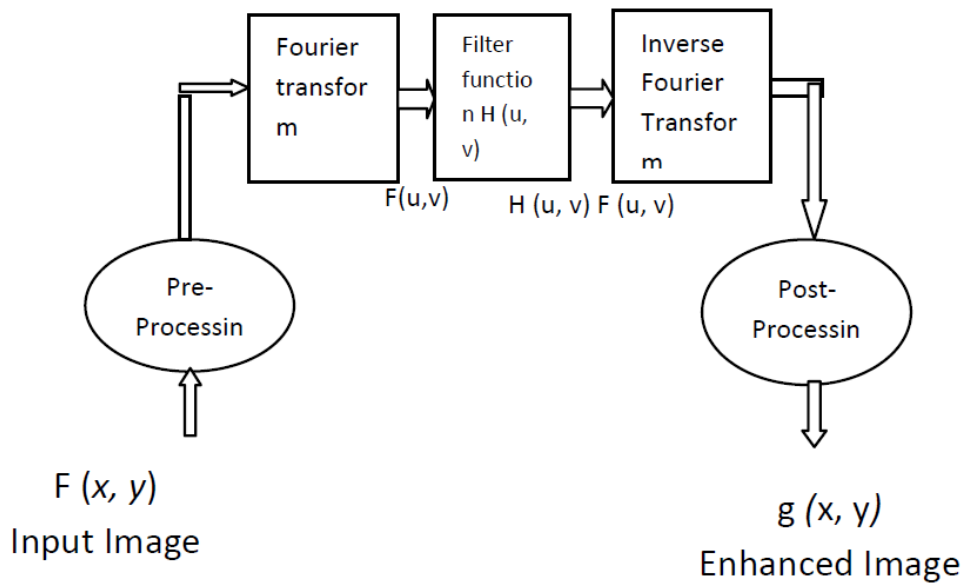

**Figure: Frequency Domain Filtering Operation**
Source: Aziz *et al*. [14]

In the paper by P. Janani *et al*. [12], image processing techniques for spatial domain are discussed. They proposed that by adding certain type of noises, one can find out which method is efficient for which kind of noise. Different types of noises like Gaussian noise, Salt and pepper noise, Speckle noise, Poisson noise and quantization noise are dealt with several filters like: 1) Median filter: This filter removes the noise while preserving the edges, so each value of the matrix is replaced by the median value. 2) Mean filter: In this filter a mask is applied on each pixel to enhance the image quality. The mask is a collection of pixels that fall under same category. 3) Weiner filter: The Weiner filter is an old approach in which the values of the pixels are optimized to reduce the noise in the signal by comparing it with desired noiseless channel. To reduce overfitting and to improve performance on test images, a large training set is required to train deep learning techniques. Hence, data augmentation is an important step before we start training the model.

|  | Gaussian Noise | Salt And Pepper Noise | Speckle Noise | Poisson Noise |
|---|---|---|---|---|
| Median Filter | 22.12 | 27.44 | 20.80 | 26.55 |
| Mean Filter | 23.15 | 24.65 | 22.89 | 20.00 |
| Wiener Filter | 24.37 | 23.17 | 24.51 | 27.02 |

**Figure: PSNR values**

Source: P. Janani *et al*. [12]

Augmentor was presented in 2017 by Bloice *et al*. [26] which introduced a software package, which can augment image data using a pipeline of various operations that can performed on them. It includes standard operations like rotation, scaling, cropping, translations, and not so common ones like shearing and perspective transformations. The fact that makes it best for machine learning is that it considers the understated condition that even after applying these transformations, the images will still have ground truth maintained. Even though deep learning architectures like YOLO detects bounding boxes by itself, it is imperative that the input image needs to be preprocessed to get good performance especially on test images with different background.

The research by Jeong *et al*. [28] suggested few ways to preprocess input images for efficient training of YOLO. The images are passed through image picker which randomly selects and crops class objects. The scale modifier then resizes the cropped version to the appropriate size with respect to the target size to be recognized. Further, the resized images are overlaid on different background images with similar size and background as the test images. Now a new annotation is created for the newly developed images, which will then be passed to YOLO for training.

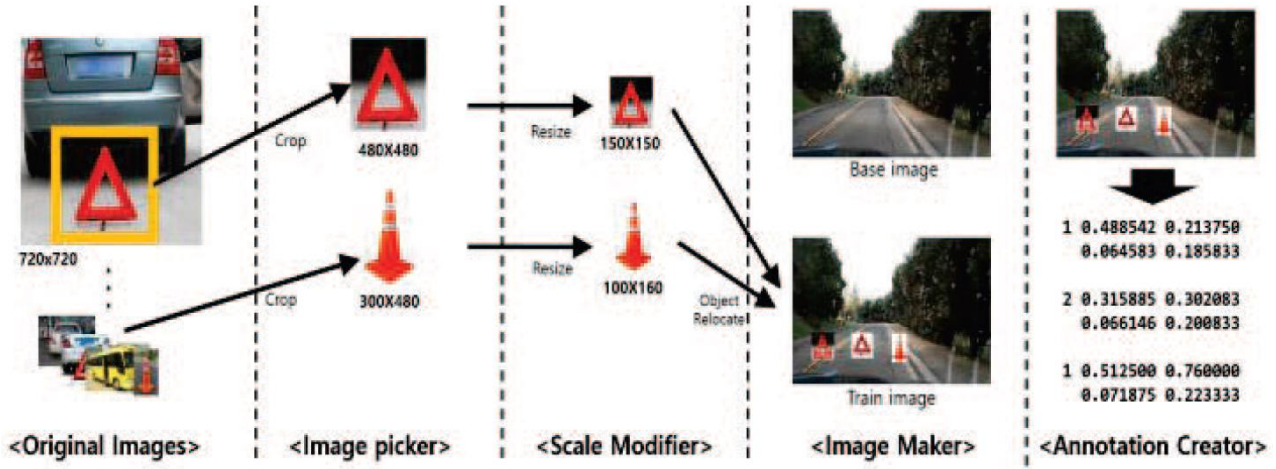

**Figure: Input Preprocessing steps for efficient training of YOLO**
Source: Jeong *et al.* [28]

## IV. IMAGE CLASSIFICATION

Once the image to be input is efficiently pre-processed, it is passed through a Classifier for recognition. Here, some of the most referred approaches are discussed.

In 2016, Faster RCNN was formulated by Ren *et al.* [17] which is a combination of fast RCNN and Region Proposal Network (RPN). RPN is a fully convolutional network that can predict bounding boxes as well as give an objectness score. Faster RCNN comprises of two modules – (i) Dense convolutional neural network used for region proposals and (ii) A Fast RCNN detector that uses the proposed regions. The computational time of faster RCNN is 10 times more than Fast RCNN as the RPNs advise the Fast RCNN detector to look at specific regions of an image and not the whole image. Also, the feature map is shared to reduce the computational time.

An extension of Faster RCNN, Mask RCNN by He *et al.* [36] is a framework that can detect as well as segment the object instance simultaneously. Along with the bounding box, category, and confidence, it also applies colored masks over the class objects. Mask R-CNN is like Faster RCNN in having two-stage network, starting with Regional Proposal Network (RPN) for bounding box prediction and ROI Pool for recognition. In Mask RCNN, they added a third output called segmentation mask, which is completely independent of the classification prediction. It was added as a third branch parallel to the second stage in Faster RCNN. Thus, with a small additional overhead to its predecessor, it was able to produce high-quality instance segmentation along with classification.

Both faster RCNN and Mash RCNN were fast, but YOLOv3 introduced by Redmon *et al.* [30] in the following year is faster and much more accurate. It is a one-stage detector which is one of the most used networks for object detection. It improved upon YOLOv2 by changing DarkNet19 with DarkNet53 which has 53 convolutional layers. They were able to achieve comparable accuracy to other detectors like SSD but in much lesser runtime.

While maintaining the accuracy and improving the speed, in 2019 Huang *et al.* [31] proposed an improved YOLOv3 architecture which can be used in real-time object recognition in industrial assembly line for electronic components. After acquiring the image dataset, they augment it by applying several image processing techniques like enhancements in contrast, brightness adjustments, smoothening effects etc. to mimic the real-time industry line images. YOLOv3 uses Darknet53, a residual model as its feature extractor. By replacing the Darknet53 with a faster detector like MobileNet, which resulted in comparable mAP and twice faster detection than YOLOv3.

| Models | YOLOV3–Mobilenet | YOLO-V3 | SSD | Faster R-CNN with Resnet 101 |
|---|---|---|---|---|
| mAP | **0.9521** | 0.9525 | 0.8904 | 0.9261 |
| Detection speed (s) | **0.0794** | 0.1569 | 0.1962 | 0.6188 |

**Figure: Comparison of mAP and Detection speed across different architectures on electronic components**
Source: Huang *et al.* [31]

As we evaluate the classifier models in 2019 which helps in accuracy vs runtime trade-off, a slight improvement in RetinaNet architecture by Yixing *et al.* [34] promises to be a great option. Faster RCNN with InceptionResNetV2 is a high-end network widely used in texture-less object recognition, which produces mAP (mean average precision) of ~36. But, at the same time, it goes through ~850 GFLOPs (Floating point operations) to achieve this result. Whereas, RetinaNet-800 achieves similar mAP with just 150 GFLOPs, which is one-fifth of Faster RCNN. Instead of a Regional Proposal Network (RPN) in Faster RCNN, RetinaNet uses a much simpler design called as Feature Pyramid Networks (FPN) which helps in lesser GFLOPs and runtime.

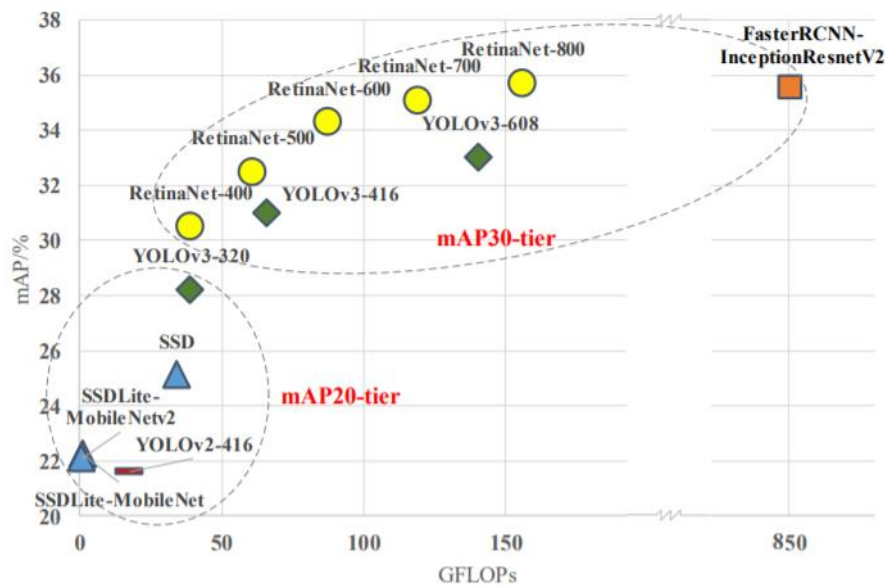

**Figure: Comparison of mAPs vs FLOPs between different architectures**
Source: Yixing *et al.* [34]

In the lightweight mAP-30 category, YOLOv3-608 achieves ~32 mAP with less than 150 GFLOPs, which is also a comparable option to RetinaNet. But on the other side, since YOLO works on SxS grid cells, it performs badly in the case of detection of smaller objects. Hence, both on mAP and GFLOPs, RetinaNet performs better than Faster RCNN and YOLOv3.

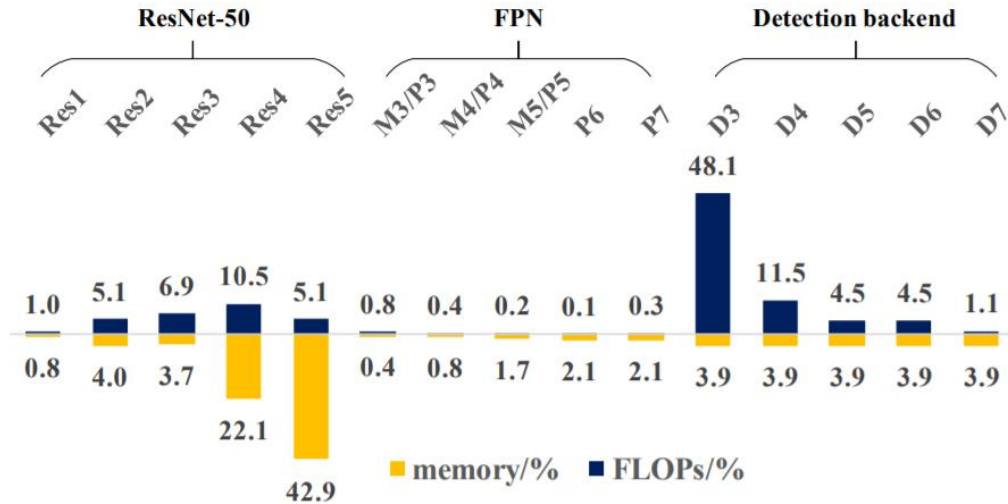

**Figure: RetinaNet layers with FLOPs and memory distribution**
Source: Yixing *et al.* [34]

Furthermore, Yixing *et al.* [34] proposed a lightweight option of RetinaNet, which achieves comparative mAP with half of GFLOPs. They achieved this by reducing the GFLOPs of D3 block in RetinaNet architecture to half of its original, which reduces the overall GFLOPs by 24%. Thus, we are left with a lightweight RetinaNet with better accuracy-runtime trade-off.

# 2. INTERNATIONAL CONFERENCE on SMART MULTIMEDIA 2018

## I. REVIEW

### 1. Spatio-temporal eye gaze data analysis to better understand team cognition

Learning about cognitive coherence of team members is imperative especially in critical industries like healthcare. It becomes much important in high precision procedures like laparoscopic surgeries. The research by Hajari *et al*. [1] introduces a method for measuring team cognition using an experimental study setup with 22 members. Two sets of features were measured: spatial features by doing an overlap analysis on eye gazes; temporal features using CRP algorithm. The combination of these features gave an understanding of how different a good performing team is from bad performing team.

### 2. Adapting texture compression to Perceptual Quality Metric for Textured 3D models

3D models which are represented using 3D mesh and 2D textures take a lot of space in storage and thus large bandwidth for streaming over networks. Hence it becomes extremely time-consuming, especially if the 3D models are not compressed into a smaller size, while loading or transferring them over networks with limited bandwidth. As we compress the models, there will be a loss of quality, but if we had some method to measure the reduction in quality, it would be easier to understand how much to compress before it goes below the desired quality. There has been a lot of work in the field of compressing the 3D mesh and measuring the quality afterwards, but very less in the field of 2D textures. The work by Kamballur *et al*. [2] focuses on measuring the loss in quality perceived after the compression of 2D textures of the 3D model.

They studied the loss using an experimental setup, where users interact and rate different models with various texture compressions. Based on the responses, they calculated Q factor, which helped in understanding the compression at which a trade-off between perceived quality and file size can be optimized.

### 3. Person Authentication by Air-writing Using 3D Sensor and Time Order Stroke Context

The research by Chiu *et al*. [3] introduces a completely new way of authenticating, by writing the signature in the air with high accuracy of >93%. The method initially converts the process of comparing the signatures as a path finding process, but in the backward direction, in the reverse order of time, which was eventually solved by a method that involved warping the time dynamically. The research also included grouping various styles of signature by same individuals into subclasses, there by removing the issue of multiplicity and finding the variance within each subclass. Using the average of this variance as verification threshold, they were able to get a high accuracy score for authentication.

### 4. Synthetic Vision Assisted Real-time Runway Detection for Infrared Aerial Images

Landing an airplane needs the pilot to accurately detect where the runway is going to be, which in turn needs to be differentiated from a touch zones which are very close to the runway. The method introduced by Liu *et al.* [4] uses synthetic vision, and then segments the infrared runway area into three: touch zone, background, and runway. Without loosing the time vs accuracy trade off, this method proposes a new way of initializing the level sets and detects the accurate runway and efficiently differentiates it from the touch zone.

### 5. Semi-Supervised Adversarial Image-to-Image Translation

Image translation creates new images, which are similar in content as source and target, but are represented in a different way. Translation between images has seen great applications in domain adaptation. For this, consistency in the domain as well as in the content are important. In domain adaptation, the labels of the unlabelled target dataset are found out using the set of labelled source dataset and few labelled target data. Eusebio *et al*. [5] introduces a Semi supervised Adversarial Translation network which is a combination of Generative Adversarial networks and Variational Autoencoders to do this task. The Variational autoencoders helps to extract the latent features of the image and GANs helps to make sure the domain and category of the translated images remain consistent. They applied this method to translate images and adapt domains for MNIST dataset.

### 6. Predict Student's Seating Distribution Based on Social Affinity⋆

To improve academic teaching strategies, it would be helpful if we understand the social network of students in class. This would allow efficient planning of teaching methods and understand their social relationships. This work by Pei *et al*. [6] focuses on two main results: Creating a Social Affinity Map and Prediction of future distribution of seating. In the experimental setup, the instructor takes the image of the class at the beginning of each class, which is then passed through a facial recognition system to identify the faces. Then the faces are localized and distance between two of them are measured using Euclidean distance. Thus, they identify the close relationships between the students using the map. Once the relationships are understood, eight neighbours for each student are identified from the map. After collecting enough data after several images captures of dataset, the seating distribution is predicted based on the assumption that close friends tend to sit closer.

### 7. A Model-Based Approach for Arrhythmia Detection and Classification

Real time detection and classification of Cardiac Arrhythmia has been of high importance in the medical industry due to the advantages it can bring by diagnosing early. Li *et al*. [7] proposed an algorithm to classify the abnormalities in heartbeat with more than 85% accuracy without loosing much on the sensitivity-specificity paradigm. Since ECG records are mixed with noises and artifacts, it is initially passed through a filter to denoise and remove the artifacts. Then each of the heartbeats are extracted using the annotations by the cardiologist. Seven features are calculated from the beats which denotes its duration, shape, and right-left ventricle depolarization. Training and testing data are handpicked to get a balanced dataset. A Support Vector Machine (SVM) and Multilayer Perceptron (MLP) are used for classification and MLP did slightly better than SVM, resulting in higher than above average accuracy and a better specificity-sensitivity difference.

### 8. Deep Reinforcement learning methods for navigational aids

Fakhri *et al*. [8] put forth a method GraphMem which makes use of MANN (memory augmented neural network) for providing a way via visual manner to people who have problem with navigation. The data is stored using both edges as well as nodes. It takes an instance at a given time and gives both context vector as well as state observation. While GraphMem was the slowest when it came to covering the entire maze, it showed significant improvement over its predecessors in round 2. The reason for this was that it used transfer learning to a great effect while shadowing its predecessor.

### 9. When smart signal processing meets smart imaging

In this paper by Wen *et al.* [9], they discuss various imaging techniques and compare them both quantitatively as well as qualitatively. Traditional techniques used legacy system constraints which limited the quality. Optionally, ways to improve the resolution by increasing it was looked at. However, new approaches like compressed sensing and High Dynamic Range (HDR) show good improvements in both image reproduction quality as well as low complexity. For HDR, PQ curve is used which provides a good visual quality to the pictures. Also, another aspect for HDR is 12-bit encoding instead of conventional 10-bit encoding. To reconstruct the signal, Nyquist theorem is applied which provides samples. Also, these samples are further reduced due to image priors, sparsity, etc. Content-aware quantization and image reshaping is applied to convert the baseline SDR images to HDR images. This not only improves the visual appeal of an image but also reduces the time complexity.

### 10. A Deep Learning Approach to Predict Crowd Behavior Based on Emotion

In this paper, Varghese *et al.* predicts the behaviour of the crowd in challenging situations by using emotions of each person. They employed 3DCNN to capture and store temporal information of the crowd's behaviour at various spatial resolutions. 3DCNN consists of pooling, convolution as well as temporal information. The algorithm uses 3DCNN and SVM to classify the emotions into various behavioural classes. The method takes video as an input, which is broken down into two non-overlapping p-frame parts – frames with crowd (happy, sad, angry, excited, neutral, etc.) and frames with no crowd knows as Nothing. It is then passed through 3 convolutional layers and 3 pooling layers. Finally, a FC layer and a SoftMax layer, which provides the output. The SoftMax layer provides classification which is then input to the SVM for further classification to provide the final output.

### 11. Adaptive Dithering Using Curved Markov Gaussian Noise in the Quantized Domain for Mapping SDR To HDR Image

Subhayan *et al*. [11] devised a method for de-banding the image when it is quantized. They generate noise (offline component) and add (online component) that noise in the image. This helps to reduce the computational expense of noise generation. They considered images in the Y-Cb-Cr color space, where Y is luma. Cb and Cr are the two chroma channels. Due to this, they only had to calculate once and save the output. They considered images with 10-bit encoding and thus has 1024 code-words with integer values in the range [0, 1023]. Due to 8-bit quantization per channel, their method's input has only 256 code-words. This noise works at a pixel level of the image, to reduce the de-banding effect, when SDR images are converted to HDR. This method produced better results as compared to gaussian noise or low-pass filter.

### 12. EREL-Net: A Remedy for Industrial Bottle Defect Detection

Traditional defect detection approaches either relied on human abilities or are automated for only a specific type of defect. Detection of defect in bottles is a challenging task as they are textureless. Patel *et al*. [12] make use of EREL approach. EREL is pre-dominantly used for defect detection. It is a two-part process – first, the defect region is detected via some edge amplification technique and in the second stage the regions are classified. Before classification, the regions are pre-processed. During the edge amplification stage, the gradient of each color channel is calculated and the maximum of the three is taken into consideration. Next, in the classification stage, the defect regions are manually labeled as '1' if there is defect and '0' for defect-free regions. This data is fed into a CNN classifier which has two convolutional layers and three dense layers. The activation function used was ReLU, except the last layer where SoftMax was used. The classifier was able to classify defective regions with an accuracy of 77%.

### 13. Adversarial Training for Dual-Stage Image Denoising Enhanced with Feature Matching

Sun *et al*. [13] proposed a method for denoising of image as well as recovering fine grained details using a three-level CNN. In the first stage they used Residual learning technique to denoise the image. The output was a denoised image with low quality visuals. The second stage makes use of Resnet with skip connection to reconstruct the image. The third and final stage is a discriminator which functions as a discriminator between cleaned images and original images. For testing, they used images from a well-known dataset. The images were first cropped to a desired pixel size (40x40). Next various samples were created by randomly translation, rotating, and scaling the images. Different noises (sigma = 30, 50 and 70,) were introduced to these images. The final output (denoised image) was obtained via the two-step denoisers.

### 14. A Flexible Method for Time-of-Flight Camera Calibration Using Random Forest?

In this method proposed by Xu *et al*., only two components are required for efficient denoising of ToF images. It takes input from Time of Flight camera and a secondary camera. The noisy images from ToF camera are calculated using random forest. The properties of this noise model are restored by distortion models. First, an infrared ray is transmitted onto a surface and back. Next, the phase difference between the transmitted ray and reflected ray is calculated and thus the distance is obtained. The training images are obtained by placing a chessboard in-front of a ToF camera which gives 150 depth images. For classification using a random forest, a binary test is performed where the output is '1' if the index of any feature in the image is greater then a predefined threshold.

### 15. A survey on vision based hand gesture

Highlighting the importance of hand gestures in artificial intelligence, the paper by Wang *et al*. [15] performed an assessment of earlier vision-oriented works done in this field. The issues with the past surveys were such that they did not provide any statistics; relation between gesture types, classification methods and practical application were not clear. Initially, they gathered data and then compared the works that was previously done for the static as well as dynamic gestures. They concluded that contact-based recognition still lacks research while vision based has wide research areas.

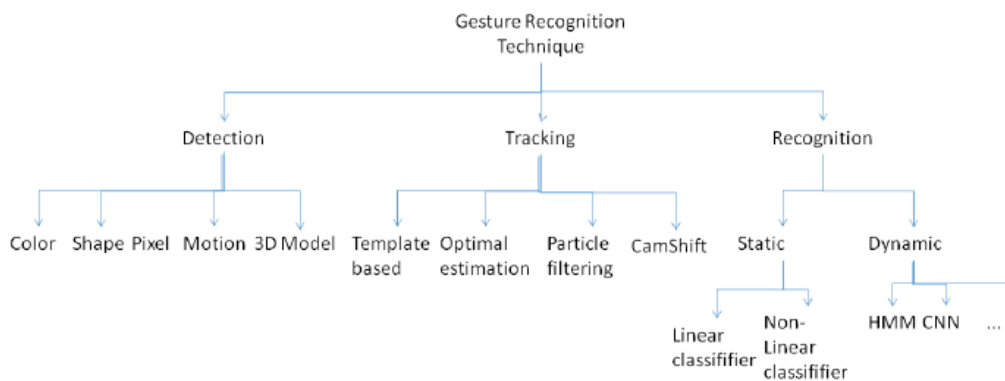

**Figure: Vision-based gesture recognition techniques**
Source: Wang *et al*. [15]

### 16. IVUS-Net: An Intravascular Ultrasound Segmentation Network

Ji *et al.* [16] describes the central problem happening in the image analysis during the diagnosis of cardiovascular diseases. As a solution, they developed an FCN (fully convolution network) architecture *IVUS-net* which was successful in segmenting the internal and external parts of the arteries after applying the post processing extraction step. Moreover, when a modern GPU is used, IVUS-net segments each frame in 0.15 seconds. FCN contains encoder and decoder network which produces embedded feature maps and refines them respectively. By skip connections the extra information is transferred from encoder to decoder. The image segmentation was enhanced by 4% to 20% when compared to the state-of-art lumen and media segmentation techniques.

### 17. Atlas-free method of periventricular hemorrhage detection from preterm infants' T1 MR Images

Detection of the hemorrhages is challenging in the presence of noise and motion effects, which is mainly due the small brain size of infants. Normally Otsu's method is used to separate out the brain from the background with noise. But it is hard to obtain an accurate brain map for them so as a solution, Subhayan *et al.* [16] proposed a different method excluding the brain maps. Earlier the brain lesion detection was segmentation based, but here they removed the segmentation of the brain tissue which made it computationally very efficient. This method detected the ventricles and focused on the white matter around ventricles rather than improving the segmentation. The area around the ventricles is marked to calculate the intensities of white matter. At the end, the outliers that were detected in uniform distribution were regarded as hemorrhages.

### 18. Automatic Computation of Fundamental Matrix Based on Voting

A new approach to reconstruct the point's geometry from several images was proposed by XinSheng *et al.* [18], to calculate the fundamental matrix. They changed the selection method and used RANSAC (random sample and consensus) framework. In this method, they used multiple low-quality fundamental matrices to combine into a better fundamental matrix. Finally, few candidates were selected based on voting to achieve a better matrix. The new matrix is the fusion of mean or fundamental summation of matrices from the initial step. Finally, few candidates are selected based on voting to achieve a better matrix. They developed a new computational method, in which fundamental matrix F is fused and voted rather than depending only on the projective distance from a pixel. This amendment makes the quality of matrix F better, with many inliers. As a conclusion RANSAC method was proved to be better than 8-point, 7-point and 5-point during the comparison study.

### 19. EREL Selection using Morphological Relation

Extremal Regions of Extremum Level (EREL) is a fresh approach for feature extraction extreme regions which is related to the segmentation problem in IVUS images. This paper by Li *et al.* [19] describes the selection process in two rounds. In the first round, the pattern of EREL regions is analyzed, which is then used to generate approximate luminal region, but 2D coefficients with highest relevance are kept. Moreover, in the other round, the compactness is calculated to fit EREL to ellipse so that it's not influenced by outliers, shadows, noises etc. EREL selection is done through several steps, firstly EREL regions are preprocessed into grayscale, next the relation between different ERELS are studied and then the EREL with highest compactness is selected. This methodology proved to be accurate when compared to the state of art.

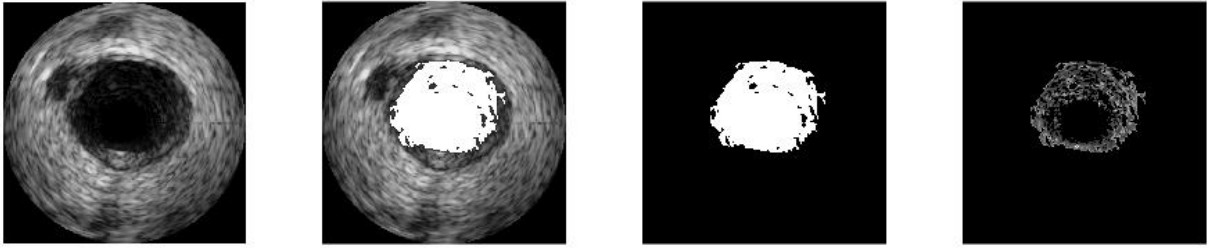

**Figure: EREL selection, segmentation, and gray scale formation from the IVUS images.**
Source: [19]

### 20. Research on path planning method of unmanned vehicle under urban road environment

The most common challenge for the autonomous driving system is to create the path in the urban environment due to presence of several unwanted obstacles. As a solution Yu Ruixing et al. [20] proposed a new method in this paper in which they considered the four wheels of the vehicle instead of whole vehicle to calculate the path. They have used the A* algorithm to find the shortest path in raster maps but to find the state of vehicle in motion they used SFS (stochastic fractal search) algorithm. Thus, by combining thee two algorithm the resulting composite algorithm helps in planning the path for autonomous vehicles.